\documentclass{article}
\usepackage{spconf,amsmath,graphicx}

\usepackage{booktabs}
\usepackage{amssymb}
\usepackage{subfigure}
\usepackage{xcolor}
\usepackage{multirow}
\usepackage{url} 
\usepackage{caption}

\title{Color Constancy in Hyperspectral Imaging via Reduced Spectral Spaces}
\name{G. Dofri Vidarsson, Liying Lu, Sabine S\"usstrunk}
\address{\'Ecole Polytechnique F\'ed\'erale de Lausanne (EPFL), Lausanne, Switzerland}
\begin{document}
%
\maketitle
\begin{abstract}
Illuminant estimation aims to infer scene illumination from image measurements despite intrinsic ambiguities between surface reflectance and lighting. Most existing methods operate on trichromatic RGB images and are therefore fundamentally limited by the restricted spectral information available. Hyperspectral imaging provides a much richer representation of scene radiance and has the potential to alleviate these ambiguities. However, its high dimensionality poses computational and statistical challenges. In this work, we systematically study the effect of spectral dimensionality and representation choice on illuminant estimation performance using hyperspectral data. We adopt the practical and effective Color-by-Correlation (CbC) framework as the estimation backbone and analyze its behavior under different spectral dimensionality reduction strategies. Our results offer practical insights into how hyperspectral information can be efficiently exploited for illuminant estimation and identify conditions under which compact spectral representations outperform conventional RGB-based approaches. The code is available at \url{https://github.com/IVRL/Reduced-Spectral-Color-Constancy}.                                                                                                                                                                                                                                                                                                                                                                                                                                                                                                                                                                                                                                                                                                                                                                                                                                                                                                                                                                                                                                                                                                                                                                                                                                                                                                                                                                                                                                                                                                                                                           
\end{abstract}
\section{Introduction}
\label{sec:intro}

Color constancy refers to the ability to perceive the color of an object as relatively constant despite changes in its lighting, e.g., the spectral composition of the scene illuminant. The human visual system achieves high color constancy through mechanisms such as chromatic adaptation. Obtaining the same robustness computationally remains a challenge in computer vision and computational photography. A fundamental component of color constancy is illuminant estimation, which seeks to infer the spectral characteristics of the scene illumination from image measurements.

Illuminant estimation is inherently ill-posed, as different combinations of surface reflectance and illumination can produce identical image observations. Most existing methods~\cite{gray_world, white_patch, gamut, cbc, barron2015convolutionalcolorconstancy, fc4, afifi2022auto, barron2017fast, afifi2021cross} operate on trichromatic RGB images, where the limited spectral information exacerbates this ambiguity and often leads to failures in complex real-world scenes.

A promising direction to improve illuminant estimation is to exploit richer spectral information. Hyperspectral imaging captures dense measurements across the visible spectrum, providing much more information about both reflectance and illumination than RGB imagery. 
Some existing methods use hyperspectral data for illuminant estimation~\cite{8546178,kaust_paper,cogo2025leveraging}. While these approaches can achieve good performance, the high dimensionality of hyperspectral data introduces substantial computational challenges, and the learned representations are often difficult to interpret. Although prior work~\cite{khan2017} has examined the effect of band count and filter configuration for classical statistical algorithms extended to hyperspectral input, it remains unclear how much spectral information is required and which aspects of the hyperspectral signal are most critical for illuminant estimation.

In this work, we investigate how hyperspectral information can be effectively leveraged for illuminant estimation by studying the role of spectral representation and dimensionality reduction. We adopt the computationally efficient Color-by-Correlation (CbC) method~\cite{cbc} as an illuminant estimation backbone and use it as a controlled testbed to evaluate different spectral representations. Using synthetically relit hyperspectral reflectance images, we compare RGB projections with compact spectral representations obtained via Principal Component Analysis (PCA)~\cite{pca}, Non-negative Matrix Factorization (NNMF)~\cite{nnmf}, and Linear Discriminant Analysis (LDA)~\cite{lda}. By varying the number of retained spectral components, we characterize the trade-offs between spectral compactness and illuminant estimation accuracy, providing insights into when reduced spectral representations can outperform conventional RGB-based approaches. Since CbC histograms are constructed for a specific sensor and illuminant set, and the learned projections likewise depend on the training distribution~\cite{nambu2003}, generalization across different sensors or datasets is outside the scope of this analysis. 

\section{Methods}
\label{sec:methodology}

\subsection{Preliminaries}



\subsubsection{Image Formation Model}
\label{sec:image_formation_model}
We assume a single global illuminant and a Lambertian model. Let $\mathbf{E}(\lambda)$ be the illuminant spectral power distribution (SPD), $\mathbf{R}(\lambda,\mathbf{x})$ the surface reflectance at pixel $\mathbf{x}$, and $\mathbf{S}_p(\lambda)$ the spectral sensitivity of channel $p\in P$. For image data uniformly sampled at wavelengths $\{\lambda_i\}_{i=1}^{d}$ with interval $\Delta\lambda$, the sensor response is
\begin{equation}
\mathbf{I}_p(\mathbf{x}) \approx \sum_{i=1}^{d} \mathbf{S}_p(\lambda_i)\,\mathbf{E}(\lambda_i)\,\mathbf{R}(\lambda_i,\mathbf{x})\,\Delta\lambda.
\label{eq:discrete_rgb_from_sampled_spectral}
\end{equation}

\subsubsection{Color by Correlation}
We adopt Color by Correlation (CbC)~\cite{cbc} as the illuminant estimation backbone and study its behavior under different spectral dimensionality reductions. In CbC, a discrete set of candidate illuminants $\mathcal{E}$ is defined, typically corresponding to commonly occurring real-world illuminants such as daylight and artificial light sources.

In our implementation, for each candidate illuminant $\mathbf{E}\in\mathcal{E}$, CbC constructs a $3$-dimensional histogram using a set of training images rendered under $\mathbf{E}$. Specifically, each RGB pixel from the training set is first projected to a 3-dimensional, intensity-invariant chromaticity by computing $L_1$-normalized RGB values.
A uniform binning scheme with $B$ bins per dimension is then applied, and each pixel contributes to a histogram bin according to its chromaticity value.

At test time, a histogram is computed from the test image in the same manner. Each candidate illuminant is scored by correlating its pre-computed histogram with the test-image histogram, and the estimated illuminant $\hat{\mathbf{E}}$ is selected as the one achieving the highest correlation score.

For further details, we refer readers to the original CbC paper~\cite{cbc}. When extending CbC to hyperspectral images, we first apply spectral projection methods to reduce the dimensionality of hyperspectral images from $d$ to $d'$. A $d'$-dimensional histogram is then constructed for each candidate illuminant. More details are provided in the appendix.

\subsection{Dimensionality Reduction Methods}
\label{sec:dimensionality_reduction_methods}

As discussed earlier, directly extending CbC or other illuminant estimation methods to hyperspectral data is nontrivial. As the spectral dimensionality increases, both computational cost and memory requirements grow significantly. We therefore explore several approaches for reducing the dimensionality of hyperspectral measurements from $d$ to $d'$ while preserving information relevant for illuminant estimation.

We evaluate several dimensionality reduction methods, including a camera RGB baseline, a Random Projection,  Principal Component Analysis (PCA), Non-negative Matrix Factorization (NNMF), and Linear Discriminant Analysis (LDA). We also consider an \emph{Illuminant PCA (Ill-PCA)} variant, in which the PCA basis is learned from illuminant spectral power distributions (SPDs) rather than from illuminated scene reflectance spectra.

PCA, NNMF and LDA are fitted on pixel-level chromaticities from training images illuminated by the candidate illuminants, treating each pixel spectrum as an independent sample.


\subsubsection{Camera RGB Projection (RGB)}
\label{sec:dim_red_rgb}
The RGB Projection serves as a baseline that simulates conventional trichromatic image acquisition. Given the camera spectral sensitivity matrix $\mathbf{S}\in\mathbb{R}^{3\times d}$, each spectrum $\mathbf{p}$ of the hyperspectral image is projected as
$\mathbf{z} = \mathbf{S}\mathbf{p}, \mathbf{z}\in\mathbb{R}^3$ .

\subsubsection{Random Projection (RAND)}
Random Projection provides a non-learned baseline for dimensionality reduction. The projection matrix $\mathbf{S}\in\mathbb{R}^{d'\times d}$ is formed by sampling entries independently from a uniform distribution on $[-1,1]$, and reduced representations are computed as 
$\mathbf{z}=\mathbf{S}\mathbf{p}$.
This method does not use any training data and serves as a reference for evaluating the role of learned spectral projections.

\subsubsection{Principal Component Analysis (PCA)}

PCA provides an unsupervised linear projection that maximizes variance in a reduced-dimensional space. Given a training set of hyperspectral images rendered under all candidate illuminants, each pixel is represented as a vector in $\mathbb{R}^d$. We perform PCA over the collection of all such pixel vectors to obtain a low-dimensional projection basis.

Given the computed global spectral mean $\boldsymbol{\mu}$ and the leading $d'$ principal components $\mathbf{U}_{d'} \in \mathbb{R}^{d \times d'}$, each spectrum $\mathbf{p}$ is projected as
$\mathbf{z} = \mathbf{U}_{d'}^\top (\mathbf{p}-\boldsymbol{\mu})$.

\subsubsection{Illuminant Principal Component Analysis (Ill-PCA)}

In Ill-PCA, the projection basis $\mathbf{U}_{d'} \in \mathbb{R}^{d \times d'}$ is learned from illuminant spectral power distributions (SPDs), rather than from illuminated scene reflectance spectra. Let $\mathcal{E} = \{ \mathbf{E}_1, \dots, \mathbf{E}_N \}$ denote a set of $N$ candidate illuminants, where each illuminant $\mathbf{E}_i \in \mathbb{R}^d$ represents a discretized SPD. We perform PCA on $\mathcal{E}$, yielding a set of $d'$ orthonormal basis vectors (in $\mathbb{R}^d$) that capture the dominant modes of variation in illuminant spectra. 

\subsubsection{Non-Negative Matrix Factorization (NNMF)}
\label{sec:dim_red_nnmf}
NNMF provides a non-negative low-rank representation of spectral data, consistent with the physical non-negative nature of reflectance, illumination, and sensor responses. Given the training matrix $\mathbf{X}\in\mathbb{R}_{\ge0}^{N\times d}$, NNMF learns non-negative factors
$\mathbf{X} \approx \mathbf{U}\mathbf{V}$,
where $\mathbf{U}\in\mathbb{R}_{\ge0}^{N\times d'}$ and $\mathbf{V}\in\mathbb{R}_{\ge0}^{d'\times d}$.

The factorization learns a non-negative basis $\mathbf{V}$ from the training data. Given a pixel spectrum $\mathbf{p}\in\mathbb{R}^d$, its reduced representation $\mathbf{z}\in\mathbb{R}_{\ge0}^{d'}$ is obtained by solving a non-negative least-squares problem with $\mathbf{V}$ fixed $\mathbf{z} = \arg\min_{\mathbf{z}\ge0} \|\mathbf{p} - \mathbf{z}\mathbf{V}\|_2^2$.

\subsubsection{Linear Discriminant Analysis (LDA)}
LDA is a supervised linear dimensionality reduction method that uses class labels to maximize discriminability in the projected space. In our setting, each pixel spectrum is associated with the illuminant label of the image from which it is sampled. LDA computes a projection matrix $\mathbf{W}\in\mathbb{R}^{d\times d'}$ that emphasizes inter-illuminant variation while suppressing intra-illuminant variability, yielding
$\mathbf{z} = \mathbf{W}^\top \mathbf{p}$.

\begin{figure}[t]
    \centering
    \includegraphics[width=0.99\linewidth]{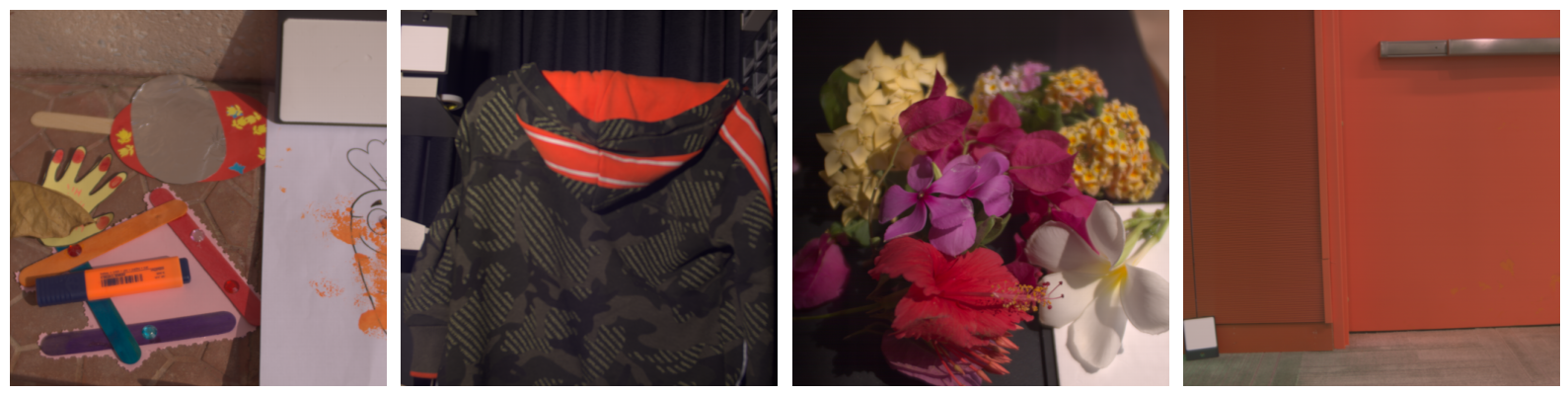}
    \caption{Examples of images from the KAUST-MIE hyperspectral reflectance image dataset. All images are rendered in sRGB under the D65 illuminant for visualization purposes.}
    \label{fig:kaust_examples}
\end{figure}

\begin{figure}[t]
    \centering
    \includegraphics[width=1\linewidth]{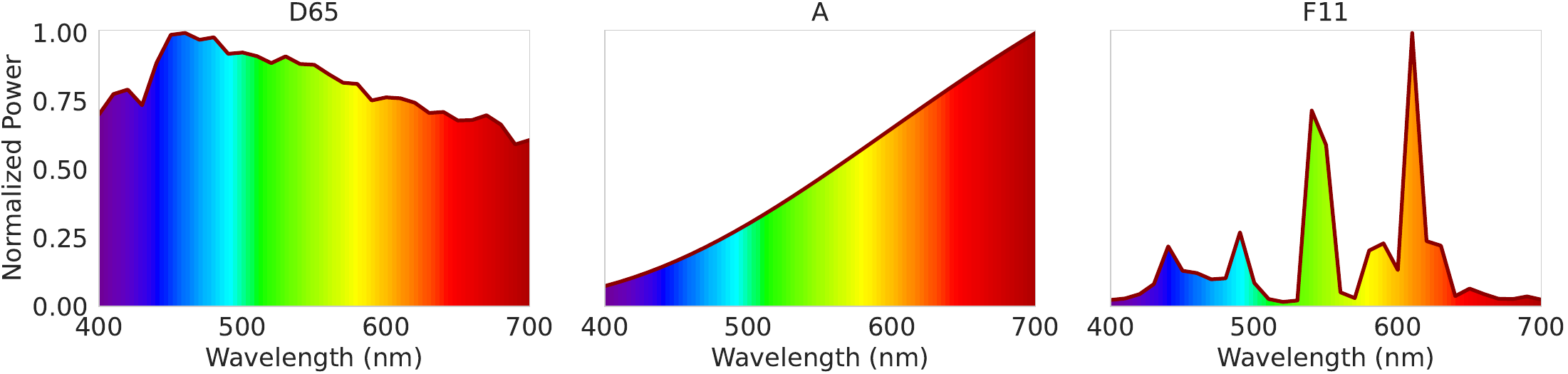}
    \caption{Example CIE standard illuminant spectral power distributions (SPDs) from the illuminant set used. Each SPD is normalized to have a maximum value of 1.}
    \label{fig:example_illuminant_spds}
\end{figure}

\section{Experimental Setup}

\begin{figure}[t]
\centering
\begin{subfigure}{}
  \includegraphics[width=0.25\linewidth]{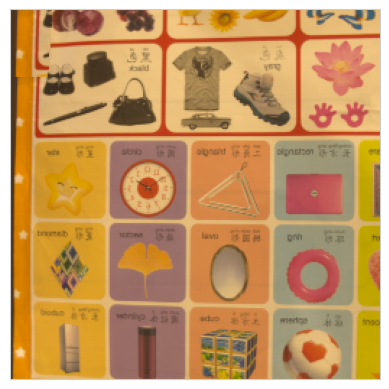}
\end{subfigure} \quad
\begin{subfigure}{}
  \includegraphics[width=0.25\linewidth]{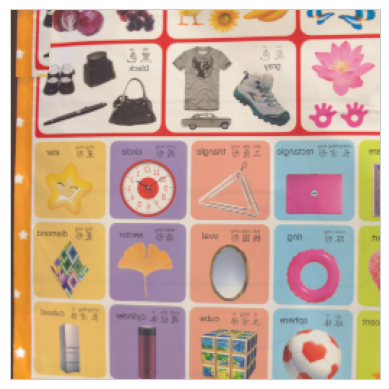}
\end{subfigure} \quad
\begin{subfigure}{}
  \includegraphics[width=0.25\linewidth]{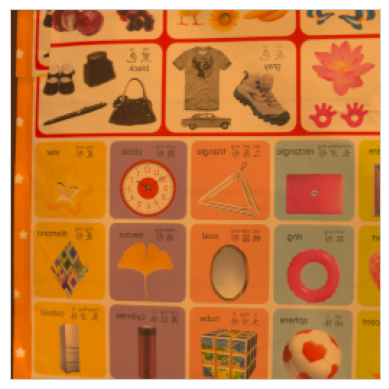}
\end{subfigure}
\caption{Synthetic relighting example showing the same scene rendered in sRGB
under three CIE standard illuminants: F2, D65, and A (left to right).}
\vspace{-0.55cm}
\label{fig:lit_images}
\end{figure}

\subsection{Dataset}

We synthesize a large number of relit spectral images (31-channel hyperspectral images) by rendering hyperspectral reflectance images under various illuminants. 


\noindent\textbf{Hyperspectral reflectance images.}
We use the KAUST Multispectral Illumination Estimation (KAUST-MIE) dataset~\cite{kaust_dataset}, which contains 409 hyperspectral images of size $512\times512$ with 31 spectral bands uniformly sampled from 400--700~nm at 10~nm intervals. The dataset covers diverse indoor and outdoor scenes (Figure~\ref{fig:kaust_examples}). Since the scene illuminant is removed using a white reference, pixel values approximate surface reflectance spectra $\mathbf{R}(\mathbf{x})$. The white reference is masked out in our experiments. These images are used both as hyperspectral inputs and to simulate RGB observations. 272 images are used as training set and 137 as test set.

\noindent\textbf{Illuminant spectral power distributions.}
We relight the reflectance images using 28 CIE Standard Illuminant SPDs from four illuminant families~\cite{cie015_2018_colorimetry}, spanning a wide range of correlated color temperatures (details can be found in the appendix). Each SPD is sampled at the same wavelengths as the hyperspectral data and represented as a 31-dimensional vector. Given an illuminant $\mathbf{E}$, the relit spectral image is computed as $\mathbf{C}(\lambda, \mathbf{x}) = \mathbf{E}(\lambda) \mathbf{R}(\lambda, \mathbf{x})$.
Figure~\ref{fig:lit_images} shows an example scene rendered under different illuminants.
The illuminants are further divided into a \emph{projection set} of 10 illuminants, used to learn dimensionality reduction transforms, and a \emph{full set} of all 28 illuminants, used for Color by Correlation histogram construction and evaluation. Details of the projection set selection are provided in the appendix.
In total, our dataset contains $409\times28=11452$ relit spectral images, 7616 for training, and 3836 for testing. Each relit spectral image has 31 channels and serves as the input to the dimensionality reduction methods. 

\noindent\textbf{Camera sensitivity functions.}
RGB images used for our RGB baseline are simulated using measured spectral sensitivity functions from three consumer cameras: Canon 300D, Nikon D90, and Sony NEX-5N from an existing camera sensitivity function database~\cite{jiang2013_camspec_paper} (visualized in the appendix). Relit spectral images are projected into the RGB space using Equation~\ref{eq:discrete_rgb_from_sampled_spectral}, enabling controlled evaluation across different illuminants and camera sensors.


\subsection{Evaluation Metric}
Performance is measured using the angular error between the estimated illuminant $\hat{\mathbf{E}}$ and the ground-truth illuminant $\mathbf{E}$
\begin{equation}
\theta(\mathbf{E}, \hat{\mathbf{E}}) = \arccos\!\left( \frac{\hat{\mathbf{E}}^\top \mathbf{E}}{\|\hat{\mathbf{E}}\|_2 \, \|\mathbf{E}\|_2} \right),
\end{equation}
which is standard in illuminant estimation and evaluates error in a scale-invariant manner by comparing illuminant directions.
Although CbC performs illuminant classification, angular error is more informative than classification accuracy as it reflects the magnitude of spectral deviation between illuminants rather than treating all misclassifications equally.

\subsection{Experimental Grid}
\label{sec:experimental_grid}

We evaluate Color by Correlation (CbC) across a grid defined by the
dimensionality reduction method, histogram resolution, and output
dimensionality. The number of histogram bins per dimension is
$B \in \{5,10,20,30\}$. For all dimensionality reduction methods except the RGB
baseline, the reduced dimensionality is varied as
$d' \in \{1,2,3,4,5\}$. The RGB baseline is fixed to $d'=3$ by construction.

The evaluated methods are Random Projection (RAND), PCA, Illuminant PCA
(Ill-PCA), NNMF, and LDA, as well as an RGB projection baseline using three
camera sensitivity functions. RAND is repeated three times with different
random seeds, and results are averaged across runs.

We additionally include a \emph{Spectral Gray World} (SGW) baseline, which estimates the illuminant as the L2-normalized mean radiance  spectrum of the test image, snapped to the nearest candidate by cosine similarity to ensure comparability with the CbC evaluation framework.

In the following sections, we primarily discuss results for $d'>1$; results for $d'=1$ are provided in the appendix.

\section{Results}

We report the mean angular illuminant estimation error in degrees for all configurations in the experimental grid. Results are averaged over the full test set. For the RGB baseline, results are averaged over the three camera sensitivity functions; individual camera results are provided in the appendix.

\begin{figure}[t]
    \centering
    \includegraphics[width=0.9\linewidth]{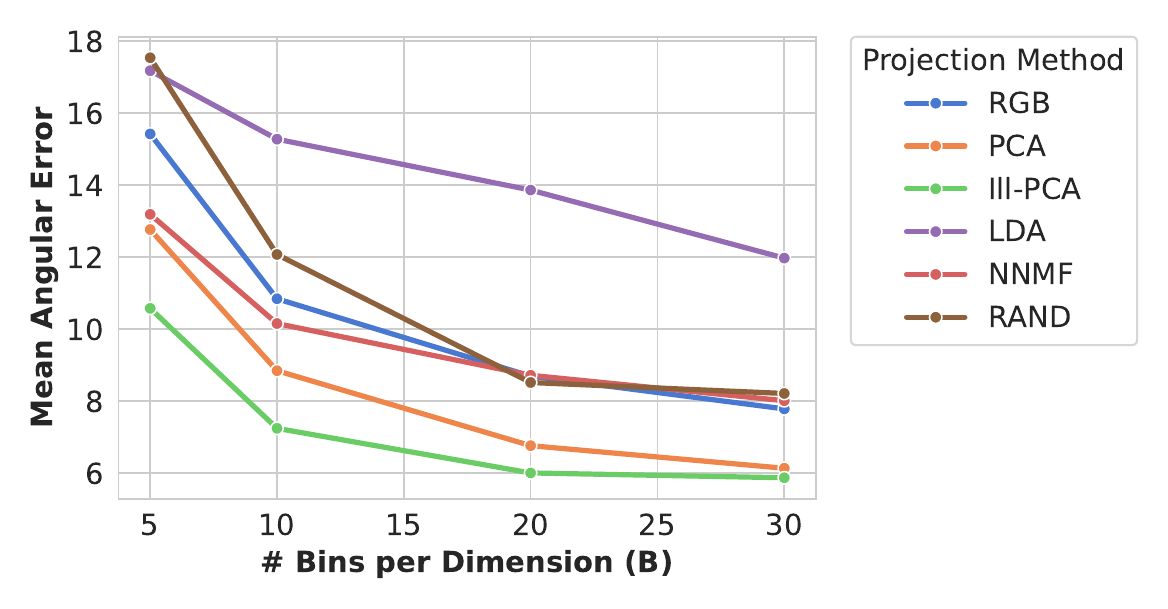}
    \caption{Mean angular illuminant estimation error (degrees) at fixed output dimension $d' = 3$ for varying histogram resolution $B$. RGB results are averaged over three camera sensitivity functions; RAND results are averaged over three random seeds.}
    
    \label{fig:3comp}
\end{figure}

\begin{table}[t]
\centering
\caption{Mean angular illuminant estimation error (degrees) at fixed histogram
resolution $B=30$ for varying output dimensionality $d'$. RGB results are
averaged over three camera sensitivity functions; RAND results are averaged
over three random seeds. SGW is dimensionality-independent and listed under
$d'=3$ for reference. Lower values indicate better performance. Best results are in bold.}

\label{tab:fixedB_dprime}
\resizebox{0.87\linewidth}{!}{
\begin{tabular}{lcccc}
\toprule
Projection Method & $d'=2$ & $d'=3$ & $d'=4$ & $d'=5$ \\
\midrule
Spectral Gray World & -- & 7.679 & -- & -- \\
RGB      & --     & 7.784  & --     & --     \\
PCA      & 10.182 & 6.136  & 3.949  & 3.129  \\
Ill-PCA & \textbf{9.314} & \textbf{5.872} & \textbf{3.737} & \textbf{2.655} \\
NNMF     & 12.981 & 8.012  & 5.852  & 4.139  \\
LDA      & 16.417 & 11.973 & 5.792  & 3.575  \\
RAND     & 14.307 & 8.213  & 4.922  & 3.382  \\
\bottomrule
\end{tabular}}
\end{table}

\subsection{Reduced Spectral Space Versus RGB}

To directly compare hyperspectral projections with conventional RGB, we fix the output dimensionality to $d'=3$, matching the dimensionality of the RGB baseline. This enables a fair comparison that isolates the effect of the spectral projection itself.

Figure~\ref{fig:3comp} reports the mean angular illuminant estimation error for each projection method at $d'=3$ as a  function of the histogram bin count $B$. Across all bin counts, PCA-based methods consistently achieve lower error than the remaining approaches. RGB performs similarly to NNMF and Random Projection, while LDA yields higher error throughout.


These trends are reflected quantitatively in Table~\ref{tab:fixedB_dprime}. At $d'=3$ and the highest tested bin count $B=30$, the average RGB angular error is $7.78^\circ$, comparable to the SGW baseline at $7.68^\circ$. In contrast, the best performing method, Illuminant PCA (Ill-PCA), achieves an error of $5.87^\circ$, corresponding to an error reduction of approximately $25\%$.

Increasing the output dimensionality further improves performance for all hyperspectral methods. At $B=30$ and $d'=5$, Ill-PCA reduces the angular error to $2.65^\circ$, representing a $66\%$ reduction relative to the RGB baseline. These results demonstrate that compact hyperspectral representations can yield substantial gains over RGB, even at low dimensionalities.



Additional error distribution statistics (median, trimean, best/worst 25\%), as well as robustness analyses under noisy conditions, are provided in the appendix.

\begin{figure}[t]
    \centering
    \includegraphics[width=0.87\linewidth]{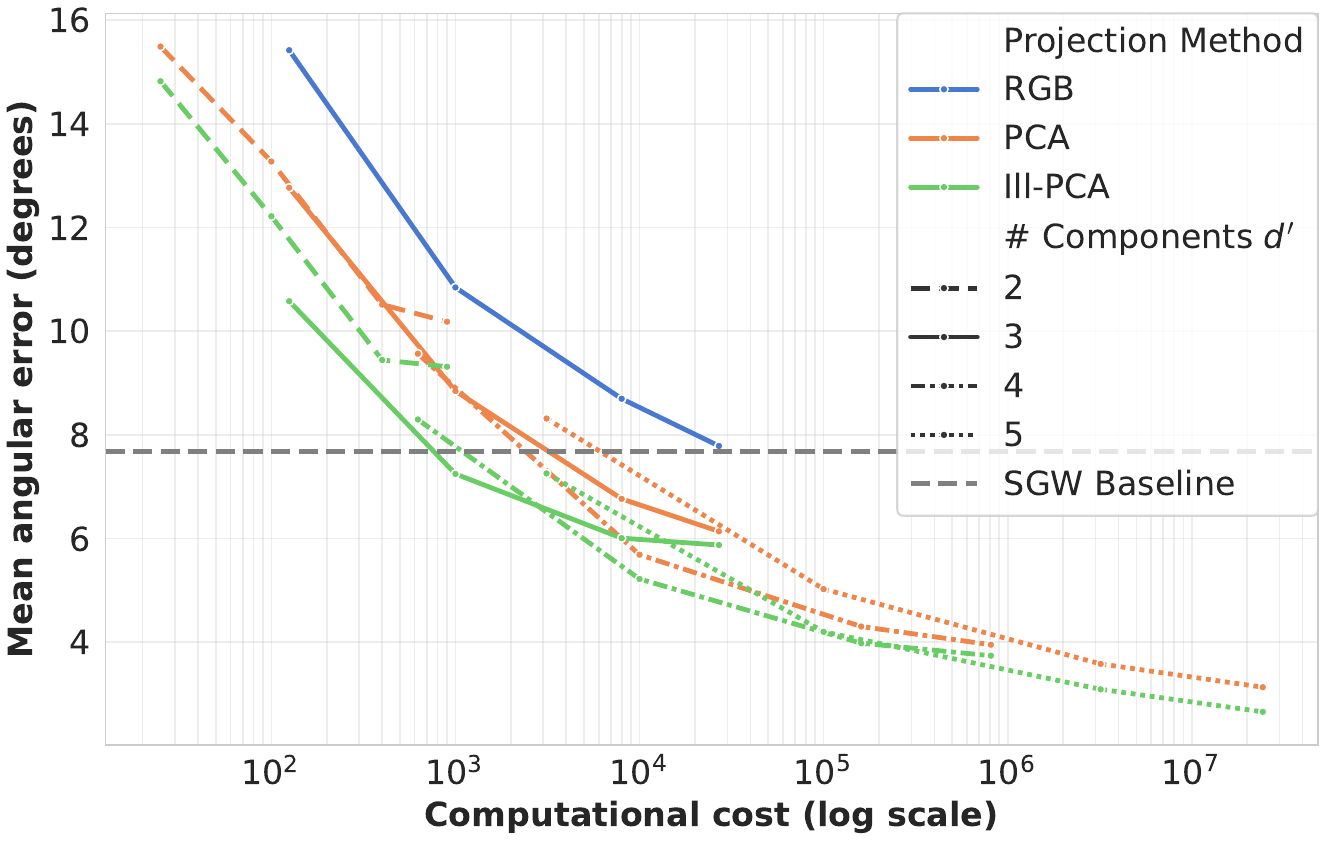}
    \caption{Trade-off between illuminant estimation error and computational cost: we show
    the mean angular illuminant estimation error as a function of computational cost (log scale). Each curve corresponds to a fixed dimensionality reduction method and output dimensionality $d'$. Lower is better for both axes. The SGW baseline is shown as a horizontal reference line.}
    \label{fig:overall_plot}
\end{figure}

\subsection{Performance and Computational Trade-offs}

Figure~\ref{fig:overall_plot} presents illuminant estimation performance as a function of computational cost, represented by the effective histogram size $B^{d'}$ in log scale within the CbC framework. Each curve corresponds to a fixed projection method and output dimensionality $d'$, evaluated across increasing bin counts $B$. RGB, PCA and Ill-PCA are shown here as they are the methods that consistently outperform RGB, and RGB acts as a baseline. 
Complete quantitative results for all projection methods, $d’$ and $B$ are provided in the appendix.

One can observe that reduced spectral representations outperform the RGB baseline for $d’ \geq 3$. While increasing $d’$ improves accuracy, it also raises computational cost. In practice, $d’ = 3$ offers a favorable trade-off between performance and efficiency.




\begin{figure}[t]
    \centering
    \includegraphics[width=1.0\linewidth]{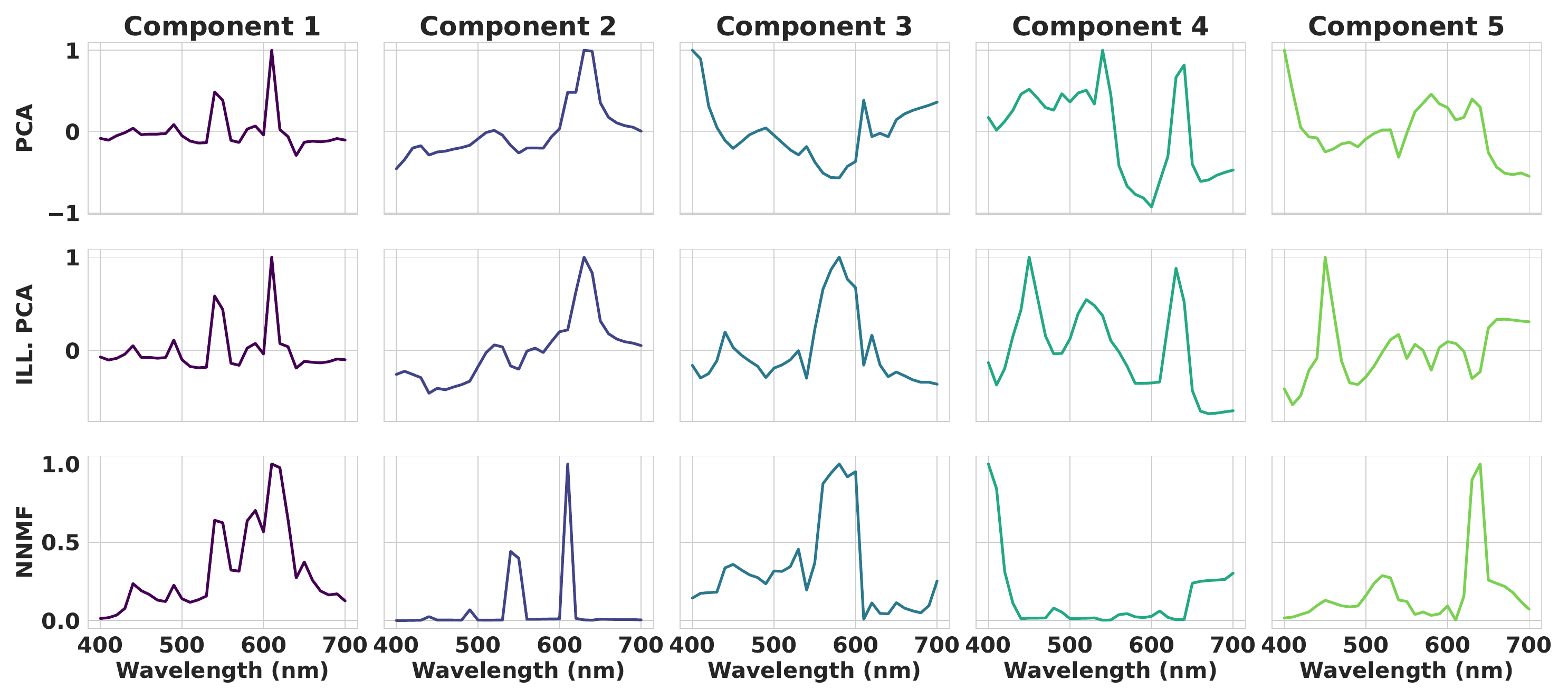}
    \caption{The learned first five basis components of PCA, Illuminant PCA (Ill-PCA), and NNMF.}
    \label{fig:projection_components_5}
\end{figure}


\subsection{Basis Interpretation}

Figure~\ref{fig:projection_components_5} shows the five learned basis
components for PCA, Illuminant PCA (Ill-PCA), and NNMF. Random Projection and LDA
are omitted, as they do not produce directly interpretable spectral bases.

The learned bases show strong qualitative similarities across methods. The first
two components of PCA and Ill-PCA are nearly identical, while the third component
captures a similar spectral structure with opposite sign. NNMF produces basis
functions with comparable patterns, including prominent peaks that coincide
with those observed in the PCA and Ill-PCA components.

Several of these peaks resemble characteristic features of fluorescent
illuminants from the F-series (Figure~\ref{fig:example_illuminant_spds}),
suggesting that the dominant components learned by all three methods are largely
driven by variations in illuminant spectra. This interpretation is further
supported by the close agreement between PCA and Ill-PCA, despite Ill-PCA being
learned solely from illuminant spectral power distributions without access to
surface reflectance data.

\subsection{Summary of Projection Method Behavior}

Across all experiments, the evaluated projection methods exhibit consistent
differences in illuminant estimation performance. PCA-based methods consistently
provide the strongest performance across dimensionalities and histogram
sizes, achieving low error even at modest output dimensions.
Notably, even when the reduced spectral dimensionality matches that of RGB ($d'=3$), PCA-based projections still outperform the RGB baseline.
Illuminant
PCA (Ill-PCA) further improves upon standard PCA, indicating that a projection
basis learned directly from illuminant spectra is particularly well suited to
the illuminant estimation task. 

NNMF, LDA and Random Projection (RAND) require higher dimensionalities ($d'\geq4$) to achieve better performance than the RGB baseline. While NNMF enforces physically meaningful non-negativity constraints, these constraints do not appear to provide a benefit for illuminant estimation under
the Color by Correlation framework. Similarly, LDA underperforms in most settings, likely due to high intra-class variability of pixel-level spectral data.
RAND performs comparably to NNMF and LDA at $d'=3$, and achieves lower error than both methods at higher output dimensionalities. This behavior suggests that, beyond a certain dimensionality, the specific shape of the projection basis is not the primary factor governing performance.

Taken together, these results indicate that for hyperspectral Color by Correlation, projection methods that provide compact yet information-rich representations are more effective than methods that impose additional constraints on the projection, particularly at low dimensionalities.

\section{Conclusions}
\label{sec:conclusion}

In this work, we studied illuminant estimation from hyperspectral images within the Color by Correlation framework, focusing on how spectral representation and dimensionality reduction affect accuracy and computational efficiency. Across PCA, Illuminant PCA (Ill-PCA), NNMF, LDA, and random projections, PCA-based methods consistently achieved the best performance, outperforming conventional RGB even when using only a small number of spectral components. Learning the projection directly from illuminant spectral power distributions further improved accuracy, indicating that the dominant structure relevant for illuminant estimation is largely driven by illuminant variability rather than surface reflectance. Overall, our results show that compact hyperspectral representations can substantially improve illuminant estimation while remaining computationally practical, motivating their use in both classical and learning-based color constancy pipelines. 
As our analysis is specific to the CbC framework, which was chosen for its simplicity and interpretability, future work could explore more advanced spectral projections and evaluate their robustness on other illuminant estimation methods, including continuous and learning-based approaches, as well as evaluating performance on real captured hyperspectral images.
\bibliographystyle{IEEEbib}
\bibliography{strings,refs}

@article{gray_world,
title = {A spatial processor model for object colour perception},
journal = {Journal of the Franklin Institute},
volume = {310},
number = {1},
pages = {1-26},
year = {1980},
issn = {0016-0032},
doi = {https://doi.org/10.1016/0016-0032(80)90058-7},
url = {https://www.sciencedirect.com/science/article/pii/0016003280900587},
author = {G. Buchsbaum},
}

@article{white_patch,
  author  = {Land, Edwin H.},
  title   = {The Retinex Theory of Color Vision},
  journal = {Scientific American},
  year    = {1977},
  volume  = {237},
  number  = {6},
  pages   = {108--128},
  month   = dec
}

@article{gamut,
  title={A novel algorithm for color constancy},
  author={David Alexander Forsyth},
  journal={International Journal of Computer Vision},
  year={1990},
  volume={5},
  pages={5-35},
  url={https://api.semanticscholar.org/CorpusID:14341320}
}

@ARTICLE{cbc,
  author={Finlayson, G.D. and Hordley, S.D. and HubeL, P.M.},
  journal={IEEE Transactions on Pattern Analysis and Machine Intelligence}, 
  title={Color by correlation: a simple, unifying framework for color constancy}, 
  year={2001},
  volume={23},
  number={11},
  pages={1209-1221},
  doi={10.1109/34.969113}}

@INPROCEEDINGS{barron2015convolutionalcolorconstancy,
  author={Barron, Jonathan T.},
  booktitle={2015 IEEE International Conference on Computer Vision (ICCV)}, 
  title={Convolutional Color Constancy}, 
  year={2015},
  volume={},
  number={},
  pages={379-387},
  doi={10.1109/ICCV.2015.51}}

@INPROCEEDINGS{jiang2013_camspec_paper,
  author={Jiang, Jun and Liu, Dengyu and Gu, Jinwei and Süsstrunk, Sabine},
  booktitle={2013 IEEE Workshop on Applications of Computer Vision (WACV)}, 
  title={What is the space of spectral sensitivity functions for digital color cameras?}, 
  year={2013},
  pages={168-179},
  doi={10.1109/WACV.2013.6475015}
}

@techreport{cie015_2018_colorimetry,
  title       = {Colorimetry, 4th Edition},
  author      = {{CIE}},
  year        = {2018},
  edition     = {4},
  institution = {Commission Internationale de l’Éclairage},
  number      = {CIE 015:2018},
  address     = {Vienna, Austria},
  doi          = {10.25039/TR.015.2018}

}

@INPROCEEDINGS{8546178,
  author={Robles-Kelly, Antonio and Wei, Ran},
  booktitle={2018 24th International Conference on Pattern Recognition (ICPR)}, 
  title={A Convolutional Neural Network for Pixelwise Illuminant Recovery in Colour and Spectral Images}, 
  year={2018},
  volume={},
  number={},
  pages={109-114},
  doi={10.1109/ICPR.2018.8546178}}

@article{article,
author = {Ahmad Khan, Haris and Thomas, Jean Baptiste and Hardeberg, Jon Yngve and Laligant, Olivier},
year = {2017},
month = {06},
pages = {1085-1098},
title = {Illuminant estimation in multispectral imaging},
volume = {34},
journal = {Journal of the Optical Society of America A},
doi = {10.1364/JOSAA.34.001085}
}

@article{khan2017,
author = {Haris Ahmad Khan and Jean-Baptiste Thomas and Jon Yngve Hardeberg and Olivier Laligant},
journal = {J. Opt. Soc. Am. A},
number = {7},
pages = {1085--1098},
publisher = {Optica Publishing Group},
title = {Illuminant estimation in multispectral imaging},
volume = {34},
month = {Jul},
year = {2017},
url = {https://opg.optica.org/josaa/abstract.cfm?URI=josaa-34-7-1085},
}

@inproceedings{nambu2003,
author = {Nambu, Satoshi and Uchiyama, Toshio and Yamaguchi, Masahiro and Haneishi, Hideaki and Ohyama, Nagaaki},
year = {2003},
month = {01},
pages = {231-235},
title = {A Method for the Unified Representation of Multispectral Images with Different Number of Bands.}
}

@inproceedings{fc4,
author = {Hu, Yuanming and Wang, Baoyuan and Lin, Stephen},
year = {2017},
month = {07},
pages = {330-339},
title = {FC4: Fully Convolutional Color Constancy with Confidence-Weighted Pooling},
doi = {10.1109/CVPR.2017.43}
}

@misc{kaust_dataset,
  author    = {Li, Yuqi and Fu, Qiang and Heidrich, Wolfgang},
  title     = {Dataset for Multispectral Illumination Estimation Using Deep Unrolling Network},
  year      = {2021},
  publisher = {KAUST Research Repository},
  doi       = {10.25781/KAUST-6930V},
  url       = {https://repository.kaust.edu.sa/items/891485b4-11d2-4dfc-a4a6-69a4912c05f1},
  urldate   = {2025-12-25}
}

@inproceedings{kaust_paper,
author = {Li, Yuqi and Fu, Qiang and Heidrich, Wolfgang},
year = {2021},
month = {10},
pages = {2652-2661},
title = {Multispectral illumination estimation using deep unrolling network},
doi = {10.1109/ICCV48922.2021.00267}
}

@article{scikit-learn,
  title={Scikit-learn: Machine Learning in {P}ython},
  author={Pedregosa, F. and Varoquaux, G. and Gramfort, A. and Michel, V.
          and Thirion, B. and Grisel, O. and Blondel, M. and Prettenhofer, P.
          and Weiss, R. and Dubourg, V. and Vanderplas, J. and Passos, A. and
          Cournapeau, D. and Brucher, M. and Perrot, M. and Duchesnay, E.},
  journal={Journal of Machine Learning Research},
  volume={12},
  pages={2825--2830},
  year={2011}
}

@article{pca,
author = { Karl   Pearson   F.R.S. },
title = {LIII. On lines and planes of closest fit to systems of points in space},
journal = {The London, Edinburgh, and Dublin Philosophical Magazine and Journal of Science},
volume = {2},
number = {11},
pages = {559-572},
year  = {1901},
publisher = {Taylor & Francis},
doi = {10.1080/14786440109462720},
}

@article{nnmf,
author = {Lee, Daniel and Seung, H.},
year = {1999},
month = {11},
pages = {788-91},
title = {Learning the Parts of Objects by Non-Negative Matrix Factorization},
volume = {401},
journal = {Nature},
doi = {10.1038/44565}
}

@inproceedings{lda,
  title={Pattern classification and scene analysis},
  author={Richard O. Duda and Peter E. Hart},
  booktitle={A Wiley-Interscience publication},
  year={1974},
  url={https://api.semanticscholar.org/CorpusID:12946615}
}

@inproceedings{afifi2022auto,
  title={Auto white-balance correction for mixed-illuminant scenes},
  author={Afifi, Mahmoud and Brubaker, Marcus A and Brown, Michael S},
  booktitle={Proceedings of the IEEE/CVF Winter Conference on Applications of Computer Vision},
  pages={1210--1219},
  year={2022}
}

@inproceedings{barron2017fast,
  title={Fast fourier color constancy},
  author={Barron, Jonathan T and Tsai, Yun-Ta},
  booktitle={Proceedings of the IEEE conference on computer vision and pattern recognition},
  pages={886--894},
  year={2017}
}

@inproceedings{afifi2021cross,
  title={Cross-camera convolutional color constancy},
  author={Afifi, Mahmoud and Barron, Jonathan T and LeGendre, Chloe and Tsai, Yun-Ta and Bleibel, Francois},
  booktitle={Proceedings of the IEEE/CVF International Conference on Computer Vision},
  pages={1981--1990},
  year={2021}
}

@article{cogo2025leveraging,
  title={Leveraging Multispectral Sensors for Color Correction in Mobile Cameras},
  author={Cogo, Luca and Buzzelli, Marco and Bianco, Simone and Vazquez-Corral, Javier and Schettini, Raimondo},
  journal={arXiv preprint arXiv:2512.08441},
  year={2025}
}
\clearpage
\setcounter{page}{1}
\appendix
\section*{Appendix}

\begin{table}[h]
\centering
\caption{CIE standard illuminants used.}
\label{tab:cie_illuminants}
\begin{tabular}{ll}
\toprule
Family & Illuminants \\
\midrule
Incandescent & A \\
Daylight (D-series) & D50, D55, D60, D65, \\
& D75, D93 \\
Fluorescent (F-series) & F1--F12 \\
LED (LED-series) & LED-B1--B5, LED-BH1, \\
& LED-RGB1, LED-V1--V2 \\
\midrule
Total &  28 \\
\bottomrule
\end{tabular}
\end{table}

\section{More implementation Details}


\subsection{Spatial Downsampling}
\label{sec:downsampling}

To reduce computation while preserving scene statistics, we perform all computations on spatially downsampled hyperspectral images. PCA and NNMF are fitted using images downsampled by a factor of 8, yielding $64\times64$ images. For LDA, we further downsample by a factor of 16, yielding $32\times32$ images, which we found sufficient for estimating stable class statistics. Color by Correlation (CbC) histogram construction and evaluation are performed at $128\times128$ resolution (downsampling by a factor of 4), and the same resolution is used for all test-time scoring.

\subsection{Ground Truth Masking}
\label{sec:masking}
The hyperspectral reflectance images include a whiteboard reference used for ground-truth calibration. This region is manually masked in all images and excluded from all subsequent processing. As a result, projection fitting, Color by Correlation histogram construction, and evaluation are performed using only pixels outside the masked region.

\subsection{CbC and Dimensionality Reduction Implementation}

In the original Color by Correlation (CbC) paper, it constructs $2$-dimensional histograms. Specifically, each RGB pixel from the training set is projected to a 2-dimensional, intensity-invariant chromaticity by computing ratios between the color channels. While in our implementation, we deviate from the original formulation by using $L_1$-normalized spectral chromaticities (instead of RGB ratio chromaticities). Both chromaticity representations are intensity-invariant and would yield similar behavior in practice.


When extending CbC to hyperspectral images, to reduce the dimensionality of the hyperspectral images from $d$ to $d'$, we use the following procedure: each spectral measurement $\mathbf{p}\in\mathbb{R}^d$ is first converted to a scale-invariant chromaticity
$\mathbf{c}=\mathbf{p}/\lVert\mathbf{p}\rVert_1$,
which removes overall intensity variation. The chromaticity is then projected to a lower-dimensional space via a dimensionality reduction transform $T$, yielding
$\mathbf{z}=T(\mathbf{c})\in\mathbb{R}^{d'}$.
A $d'$-dimensional histogram over $\mathbf{z}$ is subsequently constructed for each candidate illuminant.

PCA, NNMF, and LDA are implemented using standard scikit-learn routines (version~1.7.2)~\cite{scikit-learn}: 
\begin{itemize}\itemsep0pt \parskip0pt \parsep0pt
    \item PCA: \texttt{IncrementalPCA}
    \item NMF: \texttt{MiniBatchNMF}
    \item LDA: \texttt{LinearDiscriminantAnalysis}
\end{itemize}
Default parameters were used for all three methods.

\subsection{Candidate Illuminant Set}

\begin{figure}[t]
    \centering
    \includegraphics[width=1\linewidth]{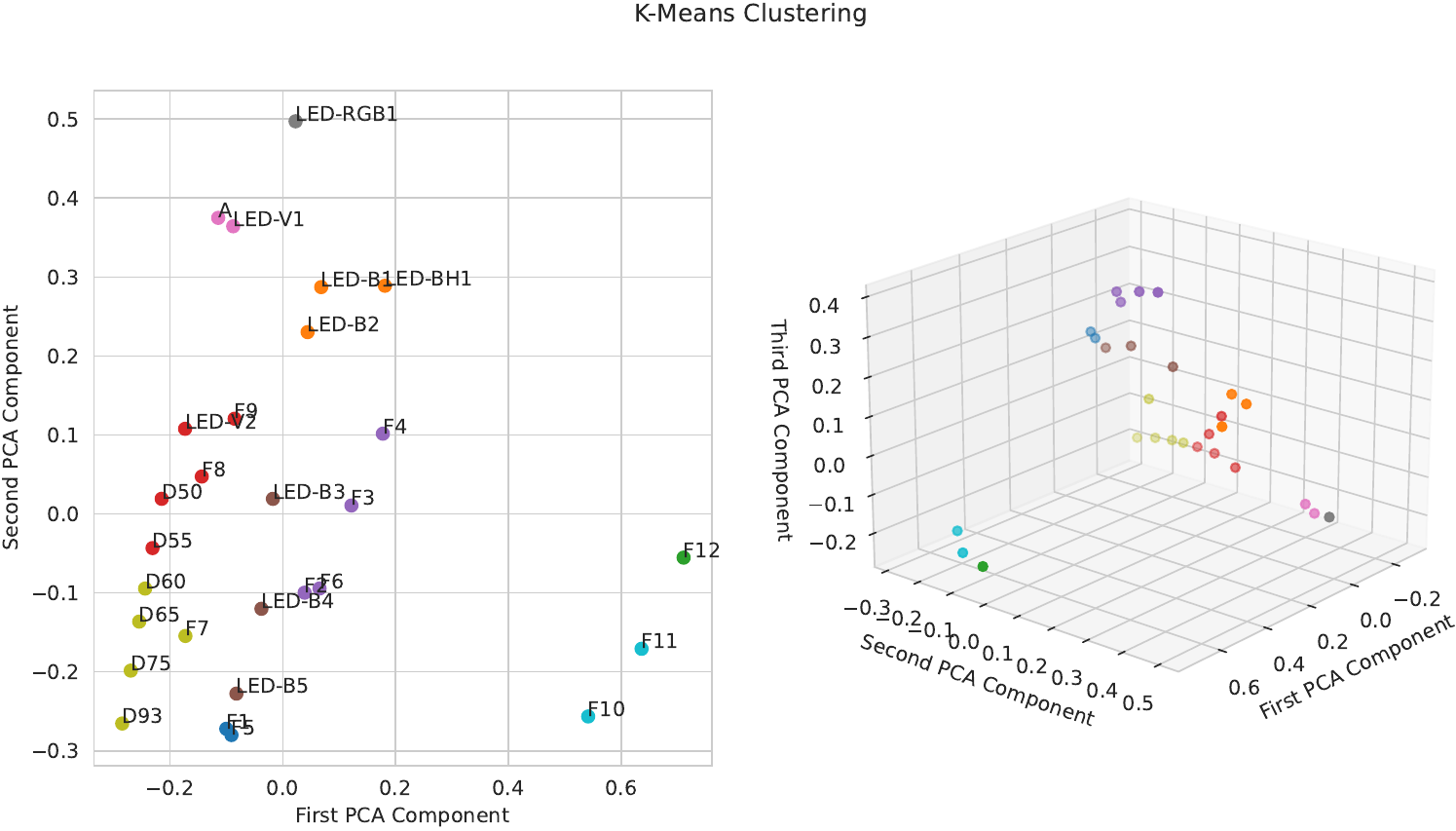}
    \caption{Illuminants clustered with $k$-means, $k=10$. Visualized on the first two and three PCA components.}
    \label{fig:k_means}
\end{figure}

\begin{figure}[t]
    \centering
    \includegraphics[width=1\linewidth]{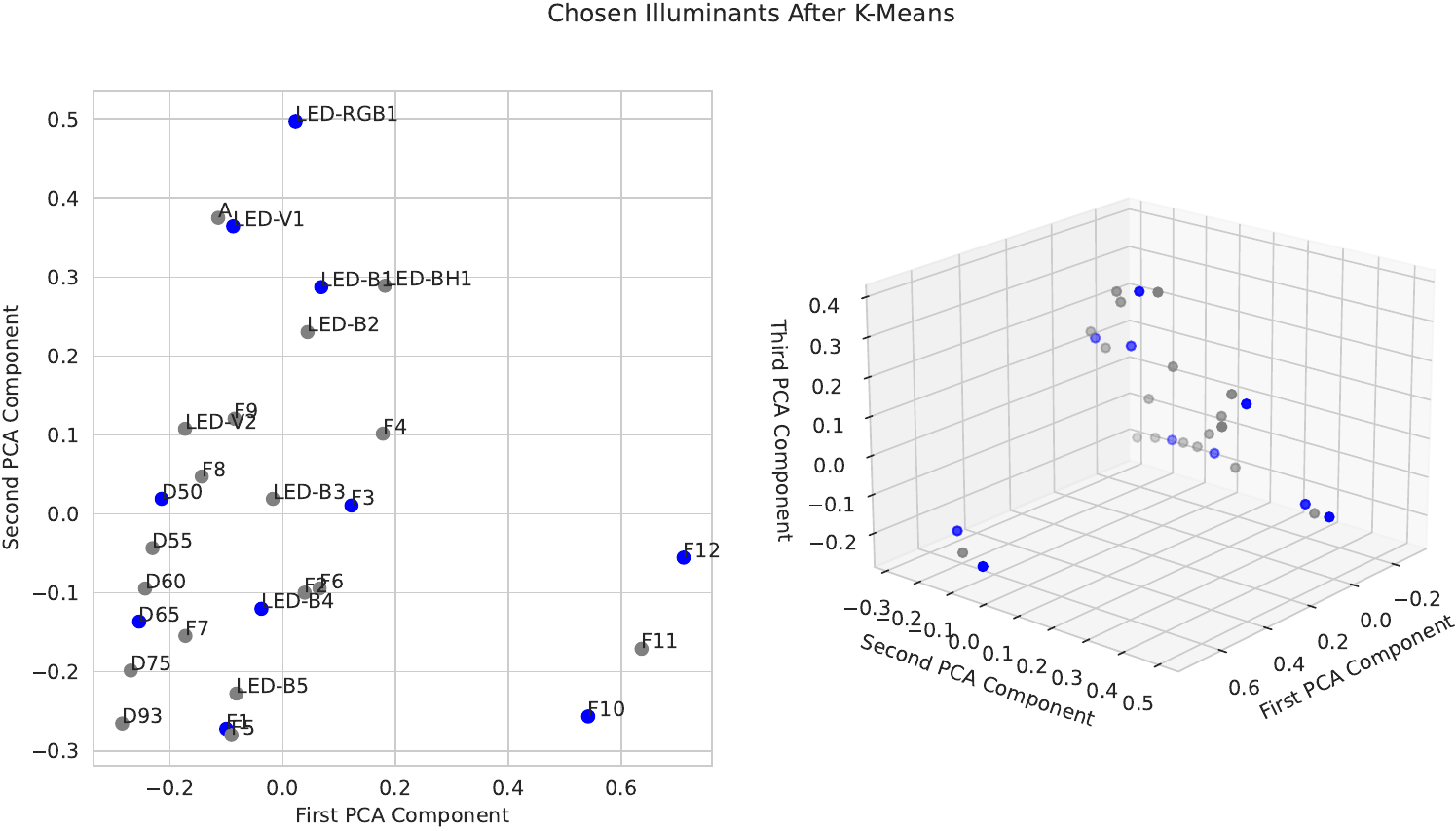}
    \caption{Chosen Projection Set. Visualized on the first two and three PCA components.}
    \label{fig:chosen_illuminants}
\end{figure}

In Table~\ref{tab:cie_illuminants}, we show the full illuminant set we used for relighting the hyperspectral images.

Furthermore, to avoid fitting our projections to skewed training data, we constructed a reduced Projection Illuminant set. The procedure for the selection is as follows. We did $k$-means clustering among the illuminants in the full illuminant set with $k=10$. Then we pick the illuminant from each cluster that is closest to that clusters centroid. Figures~\ref{fig:k_means} and \ref{fig:chosen_illuminants} illustrate the results from the $k$-means clustering and illuminant selection respectively.

\begin{figure}[t]
    \centering
    \includegraphics[width=1.0\linewidth]{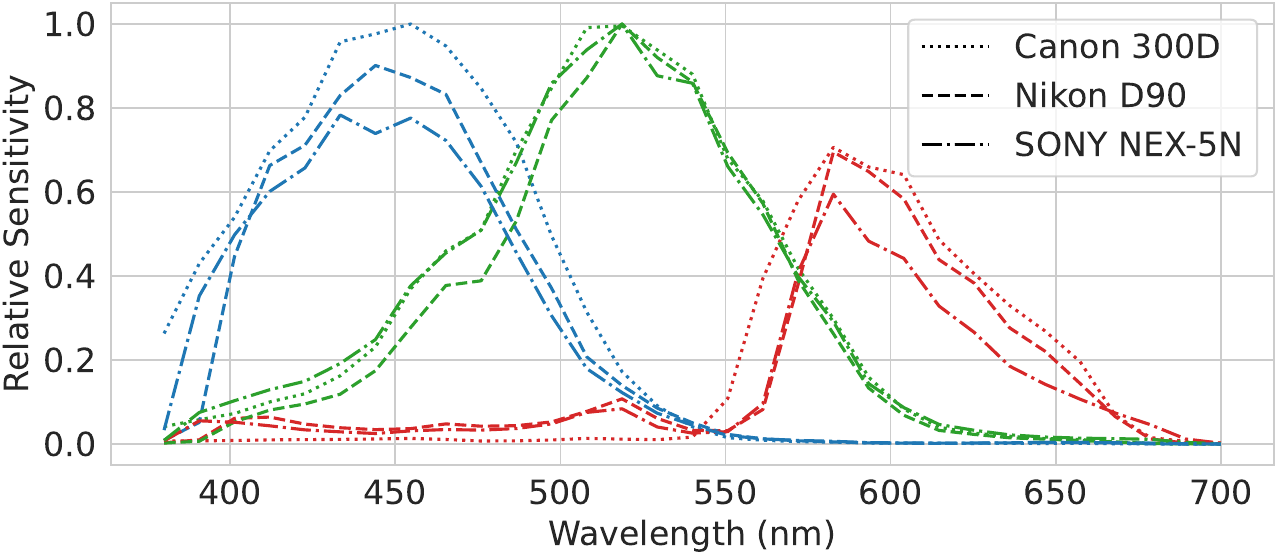}
    \caption{Spectral sensitivity functions of the three cameras represented. Each set of sensitivities is normalized so the highest value across the three functions is one.}
    \label{fig:camera_sensitivities}
\end{figure}

\subsection{Camera Sensitivity Functions}
In Figure~\ref{fig:camera_sensitivities}, we visualize the three camera sensitivity functions we used to project hyperspectral images to the RGB space.

\section{Detailed Results}

\subsection{Full Experimental Grid Results}




Tables~\ref{tab:full_results}, \ref{tab:appx_trimean}, \ref{tab:appx_q1}, and \ref{tab:appx_q4} report the complete Color by Correlation results over the full experimental grid, covering all projection methods, output dimensionalities $d'$, and histogram bin counts $B$. Results are reported as mean, trimean, best-25\%, and worst-25\% angular illuminant estimation error in degrees, respectively.

RGB results are shown only for $d'=3$ and are averaged over the three camera sensitivity functions. RAND results are averaged over three random seeds.
For each combination of $d'$ and $B$, the best (lowest) error is highlighted in bold.

\begin{figure}[t]
    \centering
    \includegraphics[width=1.0\linewidth]{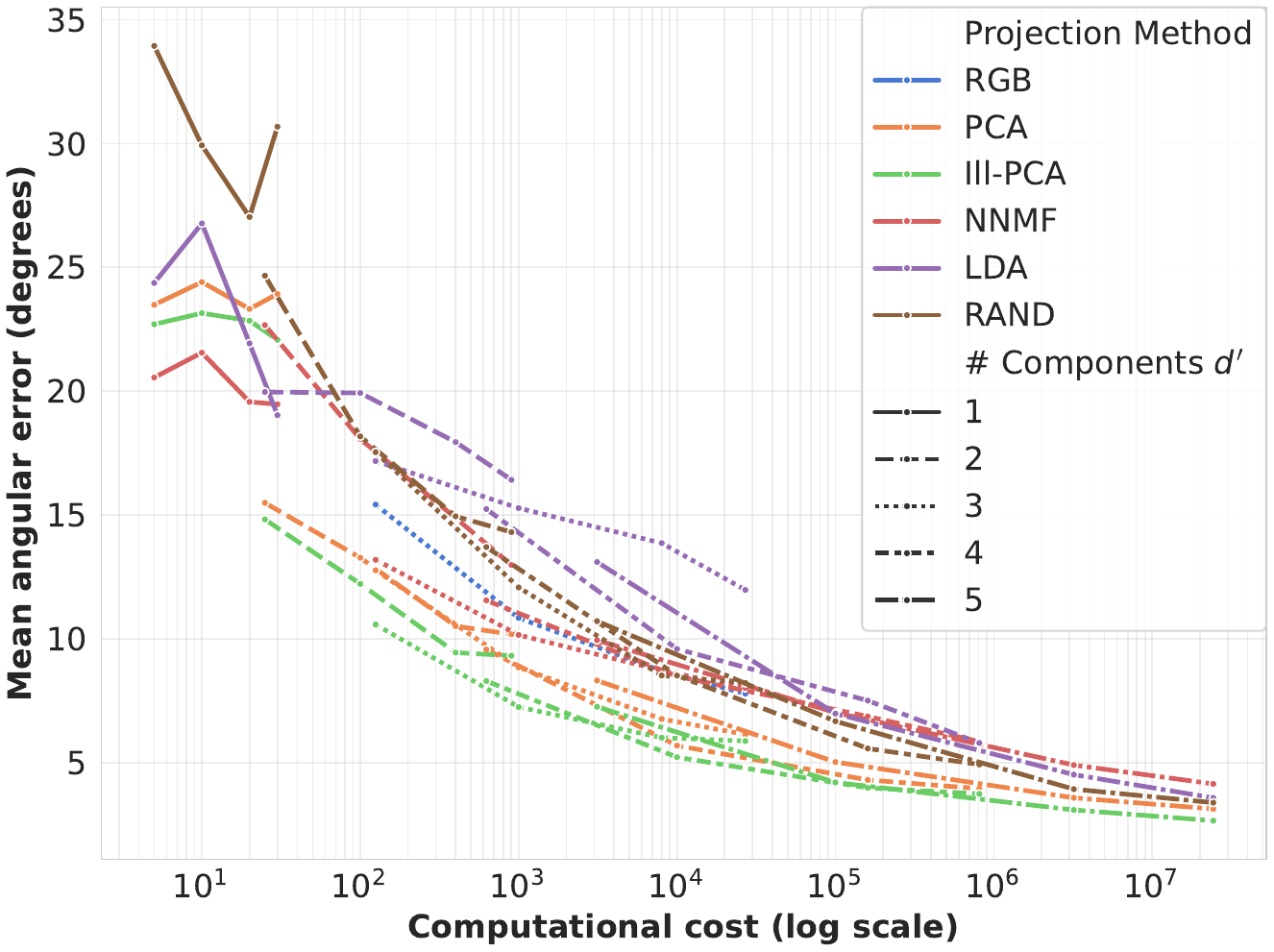}
    \caption{Mean angular illuminant estimation error as a function of
    computational cost (log scale), measured by effective histogram size
    $B^{d'}$. Each curve corresponds to a fixed projection method and output
    dimensionality $d'$. Lower values indicate better performance.}
    \label{fig:hist_size_full}
\end{figure}

Figure~\ref{fig:hist_size_full} visualizes the mean error results as a function of computational cost. Computational cost is represented by the effective histogram size $B^{d'}$ (log scale), which determines the memory and runtime requirements of the Color by Correlation framework.

\begin{table}[t]
    \centering
    \caption{Mean angular illuminant estimation error (degrees) for the three
    RGB camera models tested, evaluated at $d'=3$ for varying histogram bin
    counts $B$.}
    \label{tab:camera_results}

    \resizebox{1.0\linewidth}{!}{
    \begin{tabular}{lcccc}
    \toprule
    Camera model & $B=5$ & $B=10$ & $B=20$ & $B=30$ \\
    \midrule
    Canon 300D & 15.12 & 9.90 & 7.71 & 7.02 \\
    Nikon D90 & 15.32 & 10.97 & 8.89 & 8.01 \\
    SONY NEX-5N & 15.62 & 11.25 & 9.09 & 8.33 \\
    \bottomrule
    \end{tabular}
    }
\end{table}

\subsection{Individual RGB Camera Results}

Table~\ref{tab:camera_results} reports the illuminant estimation performance for
each individual RGB camera model. Results are shown for varying histogram bin
counts $B$ at fixed dimensionality $d'=3$ and are reported as mean angular error
in degrees.

\subsection{Individual Random Projection Results}
Table~\ref{tab:rand_runs} reports the illuminant estimation results for individual
Random Projection (RAND) runs using three different random seeds. Results are
shown for all evaluated output dimensionalities $d'$ and histogram bin counts
$B$, and are reported as mean angular error in degrees.

\section{Noise Experiment}
To assess robustness to sensor noise, we evaluate Ill-PCA at $d'=4$, $B=20$, which offers a strong balance of accuracy and computational cost, under varying levels of additive Gaussian noise. Noise is added to the radiance images (reflectance images multiplied by the L1-normalised illuminant SPD) before chromaticity normalization and scoring. The noise standard deviation is set from a target signal-to-noise ratio (SNR) as
\begin{equation}
\sigma_{\text{noise}} = \frac{\mu_{\text{signal}}}{10^{\,\text{SNR}_{\text{dB}} / 20}},
\end{equation}
where $\mu_{\text{signal}}$ is the mean radiance of the test image. Noisy values are clipped at zero to maintain physical non-negativity. Noise is sampled independently for each (image, illuminant) pair using a fixed random seed for reproducibility. We evaluate at SNR levels of $50$, $40$, $30$, $20$, and $10$~dB, alongside a clean (noise-free) reference.
Table~\ref{tab:appx_noise_illpca} reports the mean, trimean, and best/worst-25\% angular errors at each noise level.

\begin{table}[h]
\centering
\caption{Mean angular illuminant estimation error (degrees) for individual
Random Projection (RAND) runs using three different random seeds, evaluated
across output dimensionalities $d'$ and histogram bin counts $B$.}
\label{tab:rand_runs}
\resizebox{1.0\linewidth}{!}{
\begin{tabular}{llcccc}
\toprule
 & & $B=5$ & $B=10$ & $B=20$ & $B=30$ \\
\midrule
\multirow{3}{*}{${d'}=1$} & seed=42 & 33.45 & 28.77 & 23.67 & 26.06 \\
 & seed=43 & 38.95 & 34.12 & 29.34 & 28.15 \\
 & seed=44 & 30.41 & 29.19 & 34.81 & 37.83 \\
\midrule
\multirow{3}{*}{${d'}=2$} & seed=42 & 25.43 & 18.02 & 14.85 & 14.54 \\
 & seed=43 & 23.82 & 16.80 & 14.54 & 13.06 \\
 & seed=44 & 23.21 & 20.01 & 15.64 & 15.32 \\
\midrule
\multirow{3}{*}{${d'}=3$} & seed=42 & 17.09 & 10.54 & 7.37 & 6.54 \\
 & seed=43 & 17.53 & 13.85 & 9.81 & 8.46 \\
 & seed=44 & 18.88 & 14.89 & 10.68 & 9.63 \\
\midrule
\multirow{3}{*}{${d'}=4$} & seed=42 & 13.58 & 8.29 & 5.32 & 4.54 \\
 & seed=43 & 15.21 & 9.35 & 6.10 & 5.08 \\
 & seed=44 & 12.56 & 8.35 & 5.79 & 5.14 \\
\midrule
\multirow{3}{*}{${d'}=5$} & seed=42 & 10.25 & 6.56 & 3.75 & 3.10 \\
 & seed=43 & 12.35 & 7.20 & 4.41 & 3.58 \\
 & seed=44 & 10.46 & 6.49 & 3.97 & 3.47 \\
\bottomrule
\end{tabular}
}
\end{table}

\begin{table}[h]
\centering
\caption{Angular error statistics (degrees) for Ill-PCA at $d'=4$, $B=20$ across noise levels. Lower is better.}
\label{tab:appx_noise_illpca}
\begin{tabular}{lcccc}
\toprule
Noise level & Mean & Trimean & Best-25\% & Worst-25\% \\
\midrule
Clean & 4.05 & 1.46 & 0.00 & 13.63 \\
50 dB & 4.07 & 1.46 & 0.00 & 13.61 \\
40 dB & 4.28 & 1.80 & 0.00 & 14.41 \\
30 dB & 5.00 & 2.26 & 0.00 & 16.13 \\
20 dB & 6.73 & 4.39 & 0.00 & 18.79 \\
\bottomrule
\end{tabular}
\end{table}

\newpage

\begin{table}[!t]
\centering
\small
\caption{Mean angular error (degrees) for all methods across histogram bin counts $B$ and output dimensionalities $d'$. RGB is evaluated at $d'=3$ only; RGB results are averaged over three cameras and RAND results are averaged over three seeds. Best (lowest) per column is bold.}
\label{tab:full_results}
\resizebox{0.95\linewidth}{!}{
\begin{tabular}{llcccc}
\toprule
 & & $B=5$ & $B=10$ & $B=20$ & $B=30$ \\
\midrule
\multirow{5}{*}{$d'=1$} & PCA & 23.49 & 24.41 & 23.32 & 23.93 \\
 & Ill-PCA & 22.70 & 23.15 & 22.84 & 22.07 \\
 & NNMF & \textbf{20.55} & \textbf{21.55} & \textbf{19.57} & 19.47 \\
 & LDA & 24.37 & 26.77 & 21.93 & \textbf{19.03} \\
 & RAND & 33.94 & 29.93 & 27.03 & 30.68 \\
\midrule
\multirow{5}{*}{$d'=2$} & PCA & 15.49 & 13.27 & 10.51 & 10.18 \\
 & Ill-PCA & \textbf{14.82} & \textbf{12.22} & \textbf{9.44} & \textbf{9.31} \\
 & NNMF & 22.67 & 18.07 & 14.88 & 12.98 \\
 & LDA & 19.97 & 19.93 & 17.94 & 16.42 \\
 & RAND & 24.66 & 18.17 & 14.95 & 14.31 \\
\midrule
\multirow{6}{*}{$d=3$} & RGB & 15.42 & 10.84 & 8.69 & 7.78 \\
 & PCA & 12.77 & 8.85 & 6.76 & 6.14 \\
 & Ill-PCA & \textbf{10.58} & \textbf{7.24} & \textbf{6.01} & \textbf{5.87} \\
 & NNMF & 13.19 & 10.16 & 8.72 & 8.01 \\
 & LDA & 17.18 & 15.28 & 13.86 & 11.97 \\
 & RAND & 17.54 & 12.07 & 8.52 & 8.21 \\
\midrule
\multirow{5}{*}{$d'=4$} & PCA & 9.56 & 5.69 & 4.30 & 3.95 \\
 & Ill-PCA & \textbf{8.30} & \textbf{5.22} & \textbf{3.98} & \textbf{3.74} \\
 & NNMF & 11.55 & 8.49 & 6.86 & 5.85 \\
 & LDA & 15.24 & 9.59 & 7.50 & 5.79 \\
 & RAND & 13.71 & 8.52 & 5.57 & 4.92 \\
\midrule
\multirow{5}{*}{$d'=5$} & PCA & 8.31 & 5.02 & 3.58 & 3.13 \\
 & Ill-PCA & \textbf{7.25} & \textbf{4.20} & \textbf{3.09} & \textbf{2.65} \\
 & NNMF & 9.95 & 7.02 & 4.90 & 4.14 \\
 & LDA & 13.10 & 6.98 & 4.51 & 3.57 \\
 & RAND & 10.71 & 6.67 & 3.93 & 3.38 \\
\bottomrule
\end{tabular}
}
\end{table}

~
\newpage

\begin{table}[t]
\centering
\caption{Trimean angular error (degrees) across all configurations of $d'$ and $B$. RGB results are averaged over three camera sensitivity functions; RAND results are averaged over three random seeds. Best per $(d', B)$ in bold.}
\label{tab:appx_trimean}
\resizebox{\linewidth}{!}{
\begin{tabular}{llcccc}
\toprule
 & & $B=5$ & $B=10$ & $B=20$ & $B=30$ \\
\midrule
\multirow{5}{*}{$d'=1$} & PCA & 23.71 & 24.65 & 22.76 & 23.88 \\
 & Ill-PCA & 22.97 & 23.33 & 23.14 & 21.98 \\
 & NNMF & \textbf{20.27} & \textbf{21.68} & \textbf{17.91} & \textbf{17.79} \\
 & LDA & 25.66 & 29.12 & 22.47 & 19.57 \\
 & RAND & 37.42 & 30.71 & 27.53 & 33.05 \\
\midrule
\multirow{5}{*}{$d'=2$} & PCA & 15.16 & 11.95 & 8.67 & 8.46 \\
 & Ill-PCA & \textbf{14.08} & \textbf{11.23} & \textbf{7.32} & \textbf{7.26} \\
 & NNMF & 22.89 & 17.80 & 13.91 & 11.59 \\
 & LDA & 20.64 & 20.52 & 18.41 & 15.30 \\
 & RAND & 24.27 & 16.61 & 12.10 & 12.10 \\
\midrule
\multirow{6}{*}{$d'=3$} & RGB & 12.45 & 7.20 & 4.70 & 3.37 \\
 & PCA & 12.11 & 7.06 & 4.39 & 2.87 \\
 & Ill-PCA & \textbf{9.29} & \textbf{5.01} & \textbf{2.73} & \textbf{2.71} \\
 & NNMF & 11.78 & 8.43 & 6.43 & 5.71 \\
 & LDA & 18.09 & 14.61 & 11.75 & 9.09 \\
 & RAND & 15.52 & 9.85 & 5.92 & 5.23 \\
\midrule
\multirow{5}{*}{$d'=4$} & PCA & 8.43 & 2.71 & 1.80 & 1.46 \\
 & Ill-PCA & \textbf{6.38} & \textbf{2.35} & \textbf{1.46} & \textbf{1.25} \\
 & NNMF & 10.11 & 6.12 & 4.39 & 2.70 \\
 & LDA & 14.03 & 7.73 & 5.21 & 2.26 \\
 & RAND & 10.96 & 5.95 & 2.27 & 2.11 \\
\midrule
\multirow{5}{*}{$d'=5$} & PCA & 6.31 & 2.07 & 1.13 & 0.95 \\
 & Ill-PCA & \textbf{4.91} & \textbf{1.75} & \textbf{1.12} & \textbf{0.94} \\
 & NNMF & 8.91 & 5.07 & 2.07 & 1.46 \\
 & LDA & 12.37 & 4.39 & 2.01 & 1.13 \\
 & RAND & 8.72 & 3.17 & 1.53 & 1.18 \\
\bottomrule
\end{tabular}
}
\end{table}

\begin{table}[t]
\centering
\caption{Best-25\% mean angular error (degrees) across all configurations of $d'$ and $B$. RGB results are averaged over three camera sensitivity functions; RAND results are averaged over three random seeds. Best per $(d', B)$ in bold.}
\label{tab:appx_q1}
\resizebox{\linewidth}{!}{
\begin{tabular}{llcccc}
\toprule
 & & $B=5$ & $B=10$ & $B=20$ & $B=30$ \\
\midrule
\multirow{5}{*}{$d'=1$} & PCA & 4.701 & 5.268 & 3.742 & 3.540 \\
 & Ill-PCA & 3.424 & 3.945 & 3.533 & 3.476 \\
 & NNMF & \textbf{3.031} & \textbf{2.640} & \textbf{1.542} & \textbf{1.454} \\
 & LDA & 5.241 & 4.977 & 1.963 & 1.573 \\
 & RAND & 8.352 & 5.665 & 3.906 & 6.100 \\
\midrule
\multirow{5}{*}{$d'=2$} & PCA & 0.308 & 0.008 & \textbf{0.005} & 0.006 \\
 & Ill-PCA & \textbf{0.015} & \textbf{0.006} & 0.006 & 0.006 \\
 & NNMF & 2.887 & 0.765 & 0.009 & \textbf{0.004} \\
 & LDA & 2.901 & 2.693 & 1.079 & 0.013 \\
 & RAND & 2.439 & 0.146 & 0.008 & 0.006 \\
\midrule
\multirow{6}{*}{$d'=3$} & RGB & \textbf{0.006} & 0.004 & \textbf{0.000} & 0.000 \\
 & PCA & 0.007 & \textbf{0.000} & \textbf{0.000} & 0.000 \\
 & Ill-PCA & 0.007 & \textbf{0.000} & \textbf{0.000} & 0.000 \\
 & NNMF & 0.112 & 0.005 & \textbf{0.000} & 0.000 \\
 & LDA & 1.125 & 0.012 & \textbf{0.000} & 0.000 \\
 & RAND & 0.163 & 0.006 & 0.001 & 0.000 \\
\midrule
\multirow{5}{*}{$d'=4$} & PCA & \textbf{0.005} & \textbf{0.000} & 0.000 & 0.000 \\
 & Ill-PCA & 0.007 & \textbf{0.000} & 0.000 & 0.000 \\
 & NNMF & 0.005 & \textbf{0.000} & 0.000 & 0.000 \\
 & LDA & 0.014 & 0.003 & 0.000 & 0.000 \\
 & RAND & 0.006 & \textbf{0.000} & 0.000 & 0.000 \\
\midrule
\multirow{5}{*}{$d'=5$} & PCA & \textbf{0.000} & 0.000 & 0.000 & 0.000 \\
 & Ill-PCA & \textbf{0.000} & 0.000 & 0.000 & 0.000 \\
 & NNMF & 0.005 & 0.000 & 0.000 & 0.000 \\
 & LDA & 0.009 & 0.000 & 0.000 & 0.000 \\
 & RAND & 0.005 & 0.000 & 0.000 & 0.000 \\
\bottomrule
\end{tabular}
}
\end{table}

\begin{table}[t]
\centering
\caption{Worst-25\% mean angular error (degrees) across all configurations of $d'$ and $B$. RGB results are averaged over three camera sensitivity functions; RAND results are averaged over three random seeds. Best per $(d', B)$ in bold.}
\label{tab:appx_q4}
\resizebox{\linewidth}{!}{
\begin{tabular}{llcccc}
\toprule
 & & $B=5$ & $B=10$ & $B=20$ & $B=30$ \\
\midrule
\multirow{5}{*}{$d'=1$} & PCA & 41.98 & 43.68 & 44.29 & 44.70 \\
 & Ill-PCA & 41.42 & \textbf{41.88} & 42.77 & 41.34 \\
 & NNMF & \textbf{37.60} & 42.79 & \textbf{40.61} & 40.02 \\
 & LDA & 39.03 & 43.87 & 41.29 & \textbf{35.39} \\
 & RAND & 52.49 & 50.83 & 48.65 & 50.11 \\
\midrule
\multirow{5}{*}{$d'=2$} & PCA & 30.00 & 28.92 & 25.82 & 25.27 \\
 & Ill-PCA & \textbf{29.95} & \textbf{27.86} & \textbf{24.37} & \textbf{24.26} \\
 & NNMF & 41.34 & 35.31 & 30.62 & 28.94 \\
 & LDA & 34.73 & 35.22 & 33.06 & 32.00 \\
 & RAND & 46.39 & 38.99 & 35.29 & 33.81 \\
\midrule
\multirow{6}{*}{$d'=3$} & RGB & 36.66 & 30.27 & 26.68 & 24.51 \\
 & PCA & 25.90 & 22.77 & 19.65 & 18.76 \\
 & Ill-PCA & \textbf{23.45} & \textbf{20.45} & \textbf{18.57} & \textbf{18.28} \\
 & NNMF & 28.87 & 24.90 & 23.03 & 22.23 \\
 & LDA & 31.60 & 30.71 & 34.32 & 32.44 \\
 & RAND & 38.32 & 29.39 & 23.79 & 23.14 \\
\midrule
\multirow{5}{*}{$d'=4$} & PCA & 22.18 & 17.37 & 14.68 & 13.98 \\
 & Ill-PCA & \textbf{21.58} & \textbf{16.34} & \textbf{13.67} & \textbf{13.31} \\
 & NNMF & 26.09 & 22.57 & 20.21 & 18.66 \\
 & LDA & 30.68 & 24.42 & 21.57 & 17.89 \\
 & RAND & 33.22 & 23.63 & 17.30 & 16.06 \\
\midrule
\multirow{5}{*}{$d'=5$} & PCA & 21.09 & 16.03 & 12.44 & 11.31 \\
 & Ill-PCA & \textbf{20.69} & \textbf{14.06} & \textbf{11.19} & \textbf{10.11} \\
 & NNMF & 22.79 & 19.45 & 15.88 & 14.16 \\
 & LDA & 29.34 & 20.23 & 15.03 & 12.55 \\
 & RAND & 27.24 & 19.80 & 13.18 & 11.95 \\
\bottomrule
\end{tabular}
}
\end{table}
\end{document}